\documentclass[10pt,twocolumn,letterpaper]{article}

\usepackage{iccv}
\usepackage{times}
\usepackage{epsfig}
\usepackage{graphicx}
\usepackage{amsmath}
\usepackage{amssymb}
\usepackage{booktabs}
\usepackage{bbold}
\usepackage[ruled,vlined,linesnumbered]{algorithm2e}
\let\oldnl\nl
\newcommand{\nonl}{\renewcommand{\nl}{\let\nl\oldnl}}
\usepackage{amsmath}
\usepackage{color}
\usepackage{ifthen}
\usepackage{array,multirow}
\usepackage{adjustbox}
\usepackage{comment}
\usepackage{subcaption}
\usepackage{textcomp}
\usepackage{colortbl}
\definecolor{mygray}{gray}{.9}
\definecolor{gray}{rgb}{0.5,0.5,0.5} 
\definecolor{green}{rgb}{0, 0.4, 0} 
\definecolor{orange}{rgb}{1, 0.5, 0} 	
\definecolor{mahogany}{rgb}{0.75, 0.25, 0.0}
\definecolor{purple}{rgb}{0.6, 0, 0.6}
\definecolor{purple}{rgb}{0.6, 0, 0.6}
\definecolor{darkgreen}{rgb}{0, 0.4, 0.4} 
\definecolor{frenchblue}{rgb}{0.0, 0.45, 0.73}
\definecolor{what_color}{rgb}{0.7, 0.4, 0.3}

\newboolean{revising}
\setboolean{revising}{false}
\ifthenelse{\boolean{revising}}
{
	\newcommand{\ignore}[1]{}

	\newcommand{\Paul}[1]{\textcolor{frenchblue}{#1}}

} {
	\newcommand{\ignore}[1]{}

	\newcommand{\Paul}[1]{#1}

}

\usepackage[pagebackref=true,breaklinks=true,letterpaper=true,colorlinks,bookmarks=false]{hyperref}

\iccvfinalcopy 

\ificcvfinal\fi
\begin{document}

\title{Show, Adapt and Tell: Adversarial Training of Cross-domain Image Captioner}

\author{
Tseng-Hung Chen$^{\dagger}$, 
Yuan-Hong Liao$^{\dagger}$,
Ching-Yao Chuang$^{\dagger}$, 
Wan-Ting Hsu$^{\dagger}$, 
Jianlong Fu$^{\ddagger}$,
Min Sun$^{\dagger}$ \\ \\
$^{\dagger}$Department of Electrical Engineering, National Tsing Hua University, Hsinchu, Taiwan\\
$^{\ddagger}$Microsoft Research, Beijing, China\\
\small\texttt{\{tsenghung@gapp, andrewliao11@gapp, cychuang@gapp, hsuwanting@gapp, sunmin@ee\}.nthu.edu.tw}\\
\small\texttt{jianf@microsoft.com}
}

\maketitle

\definecolor{purple}{rgb}{0.6, 0, 0.6}
\newcommand{\andrew}[1]{\textcolor{purple}{#1}}

\definecolor{blue}{rgb}{0, 0.6, 0.6}
\newcommand{\james}[1]{\textcolor{blue}{#1}}

\definecolor{red}{rgb}{0.6, 0, 0}
\newcommand{\notsure}[1]{\textcolor{red}{#1}}

\vspace{-5mm}
\begin{abstract}
\vspace{-3mm}
Impressive image captioning results are achieved in domains with plenty of training image and sentence pairs (e.g., MSCOCO). However, transferring to a target domain with significant domain shifts but no paired training data (referred to as cross-domain image captioning) remains largely unexplored. We propose a novel adversarial training procedure to leverage unpaired data in the target domain. Two critic networks are introduced to guide the captioner, namely domain critic and multi-modal critic. The domain critic assesses whether the generated sentences are indistinguishable from sentences in the target domain. The multi-modal critic assesses whether an image and its generated sentence are a valid pair. During training, the critics and captioner act as adversaries -- captioner aims to generate indistinguishable sentences, whereas critics aim at distinguishing them. The assessment improves the captioner through policy gradient updates. During inference, we further propose a novel critic-based planning method to select high-quality sentences without additional supervision (e.g., tags). To evaluate, we use MSCOCO as the source domain and four other datasets (CUB-200-2011, Oxford-102, TGIF, and Flickr30k) as the target domains. Our method consistently performs well on all datasets. In particular, on CUB-200-2011, we achieve 21.8\% CIDEr-D improvement after adaptation. Utilizing critics during inference further gives another 4.5\% boost.

\end{abstract}

\vspace{-7mm}
\section{Introduction}
\begin{figure}[!t]    	
\begin{center}
\includegraphics[width=0.45\textwidth]{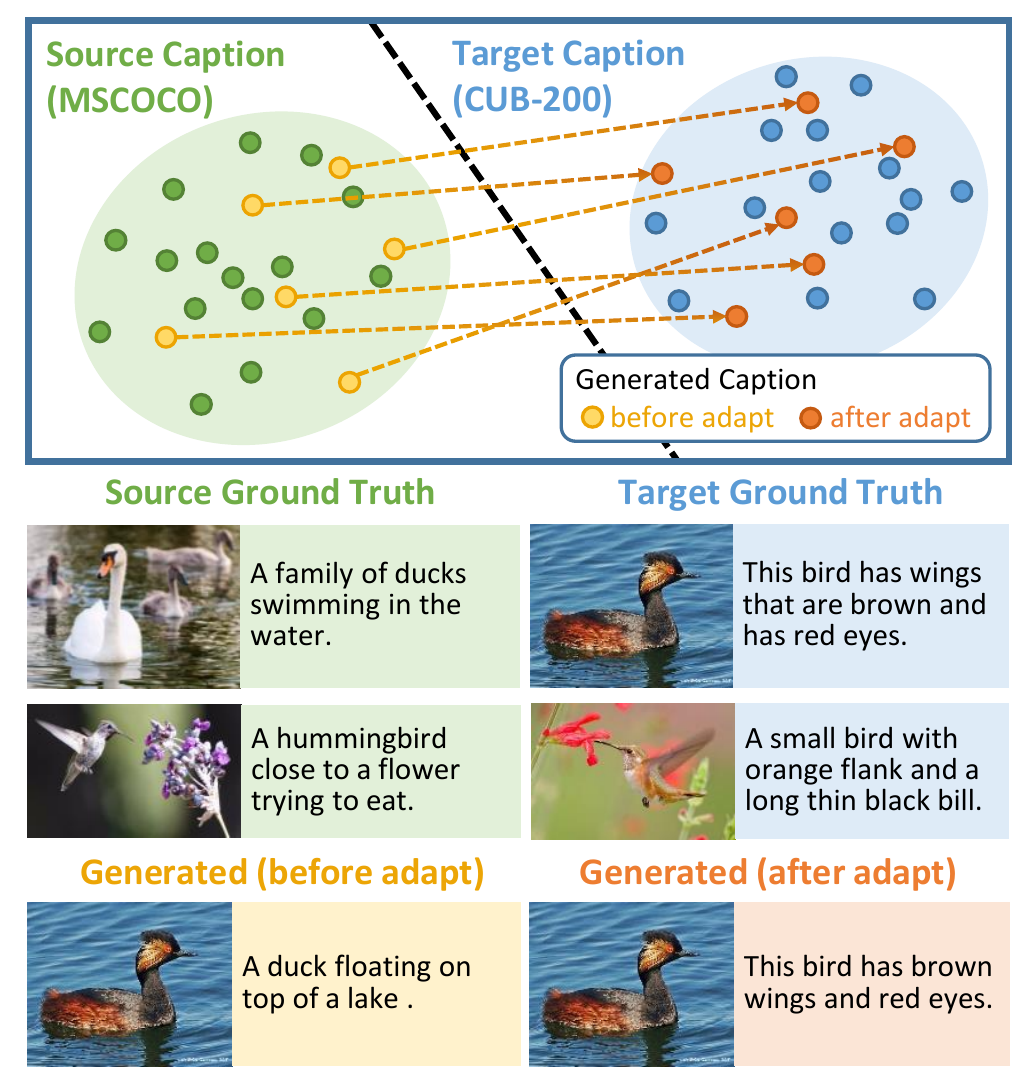}
\end{center}
\vspace{-6mm}
\caption{\small We propose a cross-domain image captioner that can adapt the sentence style from source to target domain without the need of paired image-sentence training data in the target domain. Left panel: Sentences from MSCOCO mainly focus on location, color, size of objects. Right panel: Sentences from CUB-200 describe the parts of birds in detail. Bottom panel shows our generated sentences before and after adaptation.} \label{fig:domain_example}
\vspace{-5mm}
\end{figure}


Datasets with large corpora of ``paired'' images and sentences have enabled the latest advance in image captioning. Many novel networks~\cite{donahue2015long,kiros2014unifying,karpathy2015deep,vinyals2015show} trained with these paired data have achieved impressive results under a domain-specific setting -- training and testing on the same domain.
However, the domain-specific setting creates a huge cost on collecting ``paired'' images and sentences in each domain. For real world applications, one will prefer a ``cross-domain'' captioner which is trained in a ``source" domain with paired data and generalized to other ``target" domains with very little cost (e.g., no paired data required).

Training a high-quality cross-domain captioner is challenging due to the large domain shift in both the image and sentence spaces. For instance, MSCOCO~\cite{lin2014microsoft} mostly consists of images of large scene with more object instances, whereas CUB-200-2011~\cite{wah2011caltech} (shortened as CUB-200 in the following) consists of cropped birds images. Moreover, sentences in MSCOCO typically describe location, color and size of objects, whereas sentences in CUB-200 describe parts of birds in detail (Fig.~\ref{fig:domain_example}). 
In this case, how can one expect a captioner trained on MSCOCO to describe the details of a bird on CUB-200 dataset?

A few works propose to leverage different types of unpaired data in other domains to tackle this challenge. \cite{hendricks16cvpr,venugopalan16emnlp} propose to leverage an image dataset with category labels (e.g., ImageNet~\cite{imagenet_cvpr09}) and sentences on the web (e.g., Wikipedia). However, they focus on the ability to generate words unseen in paired training data (i.e., word-level modification). Anderson et al.~\cite{AndersonFJG16a} propose to leverage image taggers at test time. However, this requires a robust cross-domain tagger. Moreover, they focus on selecting a few different words but not changing the overall style.

We propose a novel adversarial training procedure to leverage unpaired images and sentences. Two critic networks are introduced to guide the procedure, namely domain critic and multi-modal critic. The domain critic assesses whether the generated captions are indistinguishable from sentences in the target domain. The multi-modal critic assesses whether an image and its generated caption is a valid pair. During training, the critics and captioner act as adversaries -- captioner aims to generate indistinguishable captions, whereas critics aim at distinguishing them.
Since the sentence is assessed only when it is completed (e.g., cannot be assessed in a word by word fashion), we use Monte Carlo rollout to estimate the assess of each generated word. Then, we apply policy gradient~\cite{sutton1999policy} to update the network of the captioner.
Last but not least, we propose a novel critic-based planning method to take advantage of the learned critics to compensate the uncertainty of the sentence generation policy with no additional supervision (e.g., tags~\cite{AndersonFJG16a}) in testing.


To evaluate, we use MSCOCO~\cite{lin2014microsoft} as the source domain and CUB-200~\cite{wah2011caltech,reed2016learning}, Oxford-102~\cite{Nilsback08,reed2016learning}, Flickr30k~\cite{young2014image} and TGIF~\cite{Li:CVPR16} as target domains. Our method consistently performs well on all datasets. In particular, on CUB-200, we achieve 21.8\% CIDEr-D improvement after adaptation. Utilizing critic during inference further gives another 4.5\% boost. Our codes are available at \url{https://github.com/tsenghungchen/show-adapt-and-tell}.
Finally, the contributions of the paper are summarized below:
\vspace{-3mm}
\begin{itemize}
\item We propose a novel adversarial training procedure for cross-domain captioner. It utilizes critics to capture the distribution of image and sentence in the target domain.
\vspace{-3mm}
\item We propose to utilize the knowledge of critics during inference to further improve the performance.
\vspace{-3mm}
\item Our method achieves significant improvement on four publicly available datasets compared to a captioner trained only on the source domain.
\end{itemize}

\ignore{We propose a novel adversarial training procedure for cross-domain image captioning leveraging unpaired data in a target domain. Unlike previous approach which mainly attempting to modifying captions at word-level, we focus on the sentence style-level.
This takes advantages that it is easy to separately collect images and sentences in the target domain. For instance, to teach a captioner to describe birds in detail, we can collect close-up shots of birds from bird lovers' photo albums (images) and collect detailed description of birds in Wikipedia of birds (sentences).}

\ignore{However, we argue that image captioning is essentially  
Despite the huge efforts on collecting diverse image and sentence pairs in each dataset, there exists several datasets which possess in image and/or sentence with significantly different distributions. For instance, MSCOCO consists of images of large scene with more object instances, whereas CUB-200-2011 consists of cropped birds images. Hence, sentences in MSCOCO typical describe location, color and size of objects, whereas sentences in CUB-200-2011 describe parts of birds in detail (Fig. xxx).
However, most methods focus on the within domain (i.e., dataset) setting: training in one domain and testing on the same domain. 
Despite the many attempts to collect diverse}

\ignore{In order to leverage the capacity of modern deep neural networks (DNNs) for image captioning, the following setting is typically followed:  (1) supervisedly training DNNs with plenty of paired images and descriptions; (2) evaluating on other paired images and descriptions with similar source of dataset as in the training data. 
Since training and testing data are from the same source, we refer this as to closed-domain image captioning. In the past few years, many novel network architectures have been proposed [xxx] to improve the performance in closed-domain setting such as MSCOCO.}

\ignore{Image captioning is a joint vision and language modeling task, which requires datasets with image-caption pair annotations.
There are a considerable number of works~\cite{donahue2015long,kiros2014unifying,karpathy2015deep,vinyals2015show, venugopalan2015sequence,yao2015describing,zeng2016generation} that focus on supervised learning using different variants of data and encoder-decoder architectures. We discover that these captioning models perform well in a supervised setting, but perform poorly when tested on a new dataset.}

\ignore{Since the setting of annotation process for each dataset was done differently, there is a large distinction in vocabularies and phrases.
For instance, sentences in MSCOCO~\cite{lin2014microsoft} typically describe location, color and size of objects; while sentences in CUB 200-2011 dataset~\cite{wah2011caltech,reed2016learning} describe every parts of birds in detail. (see Figure.~\ref{fig:domain_shift}). 
In this paper, we propose an unsupervised domain adaptation framework for adapting between different captioning datasets.
We explore how unpaired images and sentences in the target domain dataset can help transferring to different domains.
The contributions of this paper can be summarized as follows:
\begin{itemize}
\item We propose the use of adversarial learning for domain adaptation in image captioning.
\item We introduce "Discriminator-based Planning", which makes use of the discriminator during inference stage.
\end{itemize}}
\vspace{-2mm}
\section{Related Work}
\vspace{-2mm}

\noindent\textbf{Visual description generation.}
Automatically describing visual contents is a fundamental problem in artificial intelligence that connects computer vision and natural language processing. 
Thanks to recent advances in deep neural networks and the release of several large-scale datasets such as MSCOCO~\cite{lin2014microsoft} and Flickr30k~\cite{young2014image}, many works~\cite{donahue2015long,kiros2014unifying,karpathy2015deep,vinyals2015show} have shown different levels of success on image captioning.
They typically employ a Convolutional Neural Network (CNN) for image encoding, then decoding a caption with a Recurrent Neural Network (RNN).
There have been many attempts to improve the basic encoder-decoder framework. 
The most commonly used approach is spatial attention mechanism. 
Xu et al.~\cite{xu2015show} introduce an attention model that can automatically learn where to look depending on the generated words.
Besides images, ~\cite{donahue2015long, venugopalan2015sequence,yao2015describing,zeng2016generation} apply LSTMs as video encoder to generate video descriptions.
In particular, Zeng et al.~\cite{zeng2016generation} propose a framework to jointly localize highlights in videos and generate their titles.

\noindent\textbf{Addressing exposure bias.} Recently, the issue of exposure bias~\cite{ranzato2015sequence} has been well-addressed in sequence prediction tasks. 
It happens when a model is trained to maximize the likelihood given ground truth words but follows its own predictions during test inference. 
As a result, the training process leads to error accumulation at test time. 
In order to minimize the discrepancy between training and inference, Bengio et al.~\cite{bengio2015scheduled} propose a curriculum learning strategy to gradually ignore the guidance from supervision during training.
Lamb et al.~\cite{lamb2016professor} introduce an adversarial training method as regularization between sampling mode and teacher-forced mode.
Most recently, there are plenty of works~\cite{ranzato2015sequence,bahdanau2016actor,liu2016optimization,rennie2016self} using policy gradient to directly optimize the evaluation metrics.
These methods avoid the problem of exposure bias and further improve over cross entropy methods. 
However, they cannot be applied in cross-domain captioning, since they need ground truth sentences to compute metric such as BLEU.

\noindent\textbf{Reward modeling.}
In contrast to the above works, we learn the reward function in cross-domain setting and the reward can be computed even during testing to enable our novel critic-based planning method.
Several works~\cite{hendricks2016generating,yu2016seqgan} incorporate auxiliary models as rewards.
Hendricks et al.~\cite{hendricks2016generating} minimize a discriminative loss to ensure generated sentences be class specific. 
Similar to our method, Yu et al.~\cite{yu2016seqgan} also introduce a critic to learn a reward function. However, their proposed method is for random sentence generation and not designed for domain adaptation.

\noindent\textbf{Domain adaptation.}
\Paul{Conventional DNN-based domain adaptation aim to learn a latent space that minimize the distance metrics (e.g., Maximum
Mean Discrepancy (MMD)~\cite{long2015learning} and Central Moment
Discrepancy (CMD)~\cite{zellinger2017central}) between data domains.
On the other hand, }existing adversarial domain adaptation methods use a domain classifier to learn mappings from source to target domains.
Ajakan et al.~\cite{ajakan2014domain} introduce a domain adaptation regularizer to learn the representation for sentiment analysis.
Ganin et al.~\cite{ganin2016domain} propose a gradient reversal layer for aligning the distribution of features across source and target domain. 
Hoffman et al.~\cite{hoffman2016fcns} propose an unsupervised domain adversarial method for semantic segmentations in street scenes.
\Paul{Chen et al.~\cite{chen2017no} further collect a dataset of road scene images across countries for cross-city adaptation.}
Performance improvement has been shown on sentiment analysis, image classification, person re-identification, and scene segmentation tasks.
However, we are not aware of any adversarial domain adaptation approach applied on cross-domain captioning.
\begin{figure}[!t]    	
\begin{center}
\vspace{-4mm}
\includegraphics[width=0.45\textwidth]{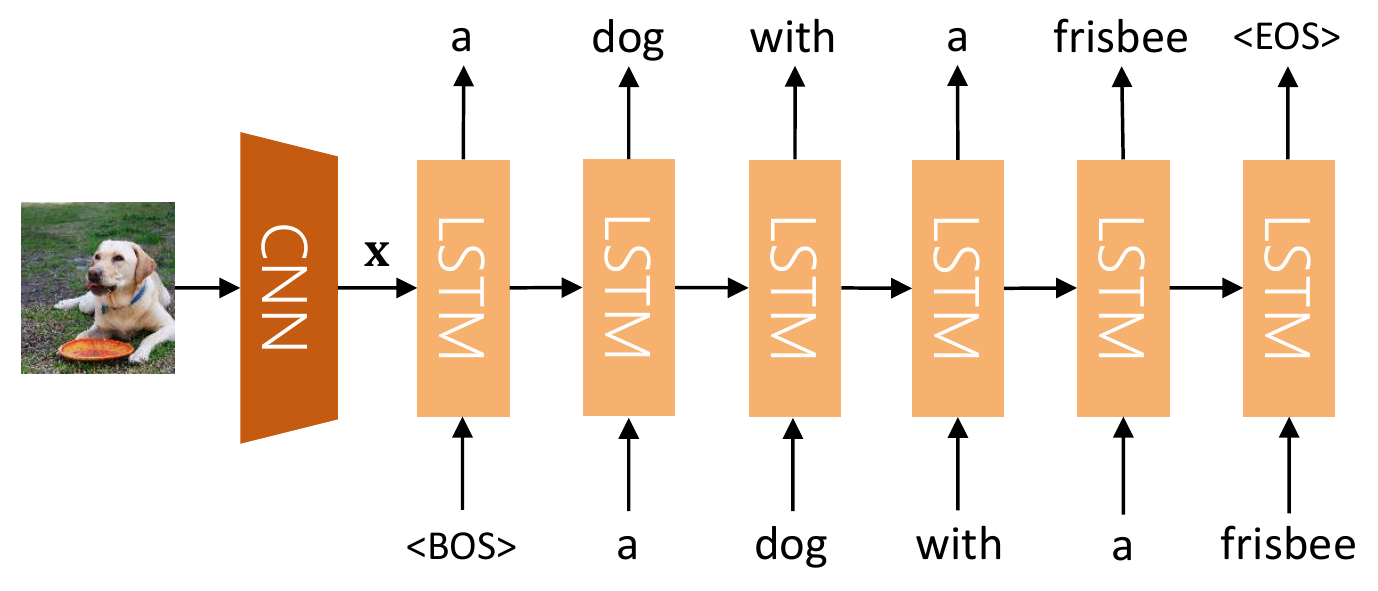}
\end{center}
\vspace{-7mm}
\caption{\small Our captioner is a standard CNN-RNN architecture~\cite{vinyals2015show}, where predicted word from previous step is serve as input of current step during inference. {\textless BOS\textgreater} and {\textless EOS\textgreater} represent the Begin-Of-Sentence and End-Of-Sentence, respectively.}\label{fig:cnn-rnn}
\vspace{-4mm}
\end{figure}

        
\section{Cross-domain Image Captioning}

We first formally define the task of cross-domain image captioning; then, give an overview of our proposed method.

\ignore{In this section, we decompose our method into five parts, including model overview, model architecture, training captioner with policy gradient, optimizing critics and discriminator-based planning.}

\noindent\textbf{Cross-domain setting.} This is a common setting where data from two domains are available.
In the source domain, we are given a set $\mathcal{P}=\{(\mathbf{x}^n,\hat{\mathbf{y}}^n)\}_n$ with paired image $\mathbf{x}^n$\footnote{We extract image representation $\mathbf{x}^n$ from CNN.} and ``ground truth" sentence $\hat{\mathbf{y}}^n$ describing $\mathbf{x}^n$. Each sentence $\hat{\mathbf{y}}=\left[\hat{y}_1,\dots,\hat{y}_t,\dots,\hat{y}_T\right]$ consists of a sequence of word $\hat{y}_t$ with length $T$.
In the target domain, we are given two separate sets of information: a set of example images $\mathcal{X}=\{\mathbf{x}^n\}_n$ and a set of example sentences $\hat{\mathcal{Y}}=\{\hat{\mathbf{y}}^n\}_n$. Note that collecting paired data $\mathcal{P}$ in the source domain is typically more costly than $\mathcal{X}$ and $\hat{\mathcal{Y}}$ in the target domain. 


\noindent\textbf{Image captioning.} For standard image captioning, the goal is to generate a sentence $\mathbf{y}$ for $\mathbf{x}$, where $\mathbf{y}$ is as similar as the ground truth sentence $\hat{\mathbf{y}}$.
For cross-domain image captioning, since the ground truth sentence of each image in $\mathcal{X}$ is not available, the goal becomes the following. For an image $\mathbf{x}\in\mathcal{X}$, we aim at generating a sentence $\mathbf{y}$ such that (1) $\mathbf{y}$ is similar to $\hat{\mathcal{Y}}$ in style, and (2) $(\mathbf{x},\mathbf{y})$ are a relevant pair similar to pairs in $\mathcal{P}$.

\noindent\textbf{Overview of our method.} To achieve the goal of cross-domain image captioning, we propose a novel method consisting of two main components. The first component is a standard CNN-RNN-based captioner (Fig.~\ref{fig:cnn-rnn}). 
However, our captioner is treated as an agent taking sequential actions (i.e, generating words).
This agent is trained using policy gradient given reward of each generated sentence. Our second component consists of two critics to provide reward. One critic assesses the similarity between $\mathbf{y}$ and $\hat{\mathcal{Y}}$ in style. The other critic assesses the relevancy between $\mathbf{x}$ and $\mathbf{y}$, given paired data $\mathcal{P}$ in the source domain as example pairs. We use both critics to compute a reward for each generated sentence $\mathbf{y}$. Both the captioner and two critics are iteratively trained using a novel adversarial training procedure.
Next, we describe the captioner and critics in detail.


\subsection{Captioner as an Agent}
At time $t$, the captioner takes an action (i.e., a word $y_t$) according to a stochastic policy $\pi_\theta(y_t|\mathbf{x},\mathbf{y}_{t-1})$, where $\mathbf{x}$ is the observed image, $\mathbf{y}_{t-1}=\left[y_1,...,y_{t-1}\right]$~\footnote{For the partial sentence starting from index $1$, we denoted it as $\mathbf{y}_{t-1}$ for simplicity. } is the generated partial sentence, and $\theta$ is the parameter of the policy. We utilize an existing CNN-RNN model~\cite{vinyals2015show} as the model of the policy. By sequentially generating each word $y_t$ from the policy $\pi_\theta(.)$ until the special End-Of-Sentence (EOS) token, a complete sentence $\mathbf{y}$ is generated.
In standard image captioning, the following total expected per-word loss $J(\theta)$ is minimized.
\vspace{-3mm}
\begin{eqnarray} \small
&&J(\theta) = \sum_{n=1}^N \sum_{t=1}^{T_n}
\mathrm{Loss}(\pi_\theta(\hat{y}_t^n|\mathbf{x}^n,\hat{\mathbf{y}}^n_{t-1}))~,\label{Eq.ori}\\
&&\mathrm{Loss}(\pi_\theta(\hat{y}_t^n|\mathbf{x}^n,\hat{\mathbf{y}}^n_{t-1})) = -\log \pi_\theta(\hat{y}_t^n|\mathbf{x}^n,\hat{\mathbf{y}}^n_{t-1})~,\nonumber
\end{eqnarray}
where $N$ is the number of images, $T_n$ is the length of the sentence $\hat{\mathbf{y}}^n$, $\mathrm{Loss}(.)$ is cross-entropy loss, and $\hat{\mathbf{y}}^n_{t-1}$ and $\hat{y}_t^n$ are ground truth partial sentence and word, respectively.
For cross-domain captioning, we do not have ground truth sentence in target domain. Hence, we introduce critics to assess the quality of the generated complete sentence $\mathbf{y}^n$. In particular, the critics compute a reward $R(\mathbf{y}^n|\mathbf{x}^n,\mathcal{Y},\mathcal{P})$ (see Sec.~\ref{critics} for details) utilizing example sentences $\mathcal{Y}$ in target domain and example paired data $\mathcal{P}$ in source domain.
Given the reward, we modify Eq.~\ref{Eq.ori} to train the agent using policy gradient.


\begin{figure*}[!t]    	
\vspace{-7mm}
\begin{center}
\includegraphics[width=1\textwidth]{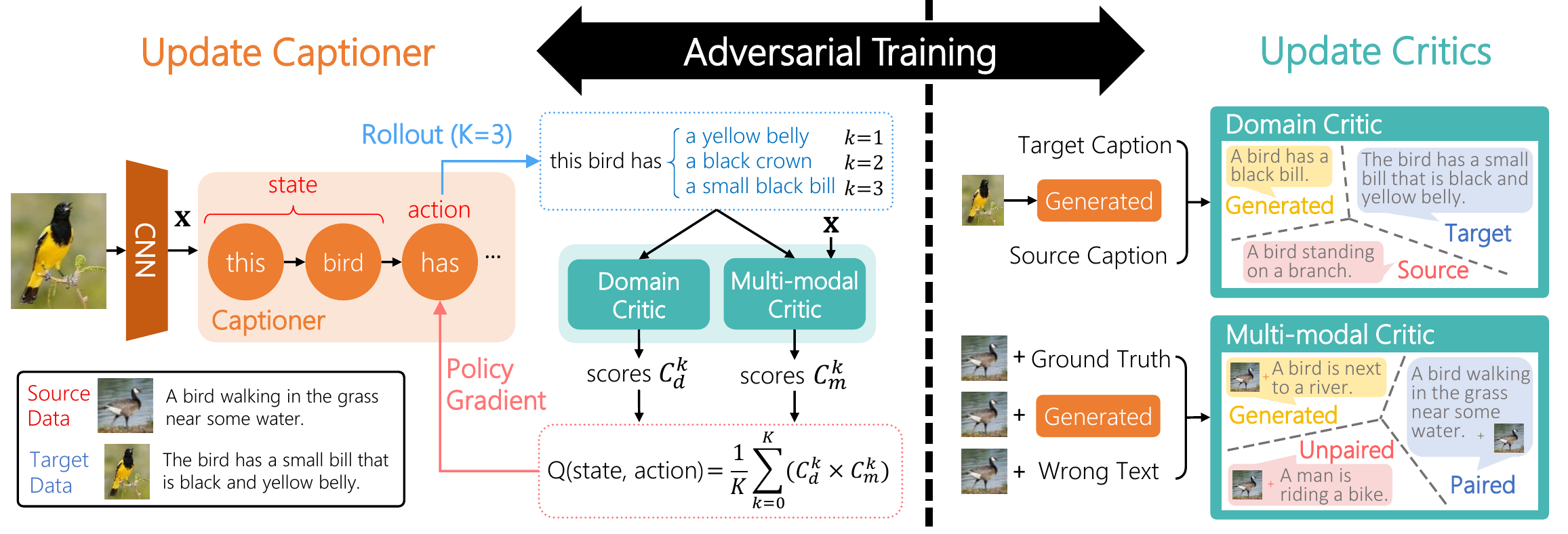}
\end{center}
\vspace{-6mm}
\caption{\small System overview. Left panel: our captioner generates a sentence condition on image representation $\mathbf{x}$. At each step, the expected reward of a newly generated word (``has") is computed from the domain and multi-modal critics using Monte Carlo rollout. We use policy gradient to update the captioner toward generating sentences with higher reward. Right panel: the critics observe sentences generated from the captioner and aim at discriminating them from the true data in target and source domains. During adversarial training, both captioner (Left) and critics (Right) are iteratively updated to achieve competing goals.
}\label{fig:model}
\vspace{-3mm}
\end{figure*}

\noindent\textbf{Policy gradient.}
The main idea of policy gradient is to replace per-word loss $\mathrm{Loss}(.)$ in Eq.~\ref{Eq.ori} with another computable term related to the state-action reward $Q(s_t,a_t)$, where the state $s_t$ is characterized by the image $\mathbf{x}$ and partial sentence $\mathbf{y}_{t-1}$ while the action $a_t$ is the current generated word $y_t$.
The state-action reward $Q((\mathbf{x},\mathbf{y}_{t-1}),y_t)$ is defined as the expected future reward:
\vspace{-1mm}
\begin{eqnarray} \small 
E_{\mathbf{y}_{(t+1):T}}\left[R(\left[\mathbf{y}_{t-1},y_t, \mathbf{y}_{(t+1):T}\right]|\mathbf{x},\mathcal{Y},\mathcal{P})\right]~.\label{Eq.Qori}
\end{eqnarray}
Note that the expectation is over the future words  $\mathbf{y}_{(t+1):T}=\left[y_{t+1},\dots,y_T\right]$ until the sentence is completed at time $T$. Hence, $Q((\mathbf{x},\mathbf{y}_{t-1}),y_t)$ takes the randomness of future words $\mathbf{y}_{(t+1):T}$ into consideration.
Given $Q((\mathbf{x},\mathbf{y}_{t-1}),y_t)$, we aim at maximizing a new objective as below,
\vspace{-6mm}\begin{eqnarray}\small 
J(\theta) = \sum_{n=1}^N J_n(\theta)~,\nonumber\\
J_n(\theta) = \sum_{t=1}^{T_n} E_{\mathbf{y}_{t}^n}\left[ \pi_\theta(y_t^n|\mathbf{x}^n,\mathbf{y}_{t-1}^n) Q((\mathbf{x}^n,\mathbf{y}_{t-1}^n),y_t^n)\right]~,\nonumber
\end{eqnarray}
where $\mathbf{y}_{t}^n=\left[\mathbf{y}_{t-1}^n,y_t^n\right]$ is a random vector instead of ground truth $\hat{\mathbf{y}}_{t}^n=\left[\hat{\mathbf{y}}_{t-1}^n,\hat{y}_t^n\right]$ as in Eq.~\ref{Eq.ori}.
However, since the spaces of $\mathbf{y}_{t}$~\footnote{We remove superscript $n$ for simplification.} is huge, we generate $M$ sentences $\{\mathbf{y}^m\}_m$ to replace expectation with empirical mean as follows,
\vspace{-1mm}
\begin{eqnarray} \small
J_n(\theta) \simeq \frac{1}{M}\sum_{m=1}^MJ_{n,m}(\theta)~,\\
\vspace{-5mm}
J_{n,m}(\theta) = \sum_{t=1}^{T_m}  \pi_\theta(y_t^m|\mathbf{x},\mathbf{y}^m_{t-1}) Q((\mathbf{x},\mathbf{y}^m_{t-1}),y_t^m)~,
\label{Eq.Jnew}
\end{eqnarray} 
where $T_m$ is the length of the generated $m^{\mathrm{th}}$ sentence.
Note that $\mathbf{y}_t^m=[
\mathbf{y}_{t-1}^m,y_t^m]$ is sampled from the current policy $\pi_\theta$ and thus computing $J_{n,m}(\theta)$ becomes tractable.
The policy gradient can be computed from Eq.~\ref{Eq.Jnew} as below,
\vspace{-1mm}
\begin{eqnarray}\small 
\triangledown_\theta J_{n,m}(\theta) = \sum_{t=1}^{T_m} \triangledown_\theta \pi_\theta(y_t^m|\mathbf{x},\mathbf{y}^m_{t-1}) Q((\mathbf{x},\mathbf{y}^m_{t-1}),y_t^m) = \nonumber\\
\vspace{-3mm}
\sum_{t=1}^{T_m} \pi_\theta(y_t^m|\mathbf{x},\mathbf{y}^m_{t-1}) \triangledown_\theta \log \pi_\theta(y_t^m|\mathbf{x},\mathbf{y}^m_{t-1}) Q((\mathbf{x},\mathbf{y}^m_{t-1}),y_t^m)~,\nonumber
\end{eqnarray}
and the total gradient is
\begin{eqnarray} \small
\triangledown_\theta J(\theta) \simeq \frac{1}{M}\sum_{n=1}^N\sum_{m=1}^M \triangledown_\theta J_{n,m}(\theta)~.\label{Eq.dJtotal}
\end{eqnarray} 
We apply stochastic optimization with policy gradient to update model parameter $\theta$.
Next we describe how to estimate the state-action reward $Q((\mathbf{x},\mathbf{y}_{t-1}),y_t)$.

\noindent\textbf{Estimating $Q$.}
Since the space of $\mathbf{y}_{(t+1):T}$ in Eq.~\ref{Eq.Qori} is also huge, we use Monte Carlo rollout to replace expectation with empirical mean as below,
\vspace{-1mm}
\begin{eqnarray} \small 
Q((\mathbf{x},\mathbf{y}_{t-1}),y_t) \simeq \nonumber\\
\frac{1}{K}\sum_{k=1}^K R(\left[\mathbf{y}_{t-1},y_t, \mathbf{y}^k_{(t+1):T_k}\right]|\mathbf{x},\mathcal{Y},\mathcal{P})~,\label{Eq.RO}
\end{eqnarray}
where $\{\mathbf{y}^k_{(t+1):T_k}\}_k$ are generated future words, and we sample $K$ complete sentences following policy $\pi_\theta$.
Next, we introduce the critics for computing the reward $R(\cdot)$.


\subsection{Critics}\label{critics}

For cross-domain image captioning, a good caption needs to satisfy two criteria: 
$(1)$ the generated sentence resembles the sentence drawn from the target domain. 
$(2)$ the generated sentence is relevant to the input image. 
The critics follow these two rules to assign reward to each generated sentence. We introduce the \textit{domain critic} and \textit{multi-modal critic} below.

\noindent\textbf{Domain critic.} 
In order to address the domain shift in sentence space, 
we train a Domain Critic (DC) to classify sentences as ``source" domain, ``target" domain, or ``generated" ones.
The DC model consists of an encoder and a classifier. 
A sentence $\mathbf{y}$ is first encoded by CNN~\cite{kim2014convolutional} with highway connection~\cite{kim2015character} into a sentence representation.
Then, we pass the representation through a fully connected layer and a softmax layer to generate probability ${C}_d(l|\mathbf{y})$, where $l\in\{source, target, generated\}$.
Note that the scalar probability ${C}_d(target|\mathbf{y})$ indicates how likely the sentence $\mathbf{y}$ is from the target domain. 

\noindent\textbf{Multi-modal critic.} 
In order to check the relevance between a sentence $\mathbf{y}$ and an image $\mathbf{x}$, we propose a Multi-modal Critic (MC) to classify $(\mathbf{x},\mathbf{y})$ as ``paired", ``unpaired", or ``generated" data.
The model of MC consists of multi-modal encoders, modality fusion layer, and a classifier as below,
\vspace{-1mm}
\begin{eqnarray} \small
\mathbf{c}=\mathrm{LSTM_{\rho}}(\mathbf{y})~,\label{Eq.1}\\
f=\tanh(W_x \cdot \mathbf{x}+b_x) \odot \tanh(W_c \cdot \mathbf{c}+b_c)~,\label{Eq.2}\\
C_m=\mathrm{softmax}(W_m \cdot f+b_m)~,\label{Eq.3}
\end{eqnarray}
where $\rho,W_x,b_x,W_c,b_c,W_m,b_m$ are parameters to be learned, $\odot$ denotes element-wise multiplication, and $C_m$ is the probabilities over three classes: paired, unpaired, and generated data.
In Eq.~\ref{Eq.1}, the sentence $\mathbf{y}$ is encoded by an LSTM-based sentence encoder. Then, in Eq.~\ref{Eq.2}, the encoded image $\mathbf{x}$ and sentence $\mathbf{c}$ representations are fused via element-wise multiplication similar to \cite{antol2015vqa}.
Finally, in Eq.~\ref{Eq.3}, the fused representation is forwarded through a fully connected layer and a softmax layer to generate probability $C_m(l|\mathbf{x},\mathbf{y})$, where $l\in\{paired,unpaired,generated\}$.
The scalar probability $C_m(paired|\mathbf{x},\mathbf{y})$ indicates how a generated caption $\mathbf{y}$ is relevant to an image $\mathbf{x}$.
\Paul{
Please see Supplementary for the intuition and empirical studies of the design choices in DC and MC.
}

\vspace{-1mm}
\noindent\textbf{Sentence reward.}
We define the reward $R(\mathbf{y}|.)=C_d(target|.) \cdot C_m(paired|.)$. This ensures a sentence receives a high reward only when (1) DC believes the sentence is from the target domain, and (2) MC believes the sentence is relevant to the image.

\vspace{-1mm}
\noindent\textbf{Training critics.}
We introduce the training objective of DC and MC below.
For DC, the goal is to classify a sentence into source, target, and generated data. This can be formulated as a supervised classification training objective as follows,
\vspace{-4mm}
\begin{equation} \small 
\begin{split}
\mathcal{L}_{d}(\phi)=-\sum_{n=1}^{N}\log C_d(l^n|\mathbf{y}^n;\phi)\\
l^n=
\begin{cases}
source & \text{if}\ \mathbf{y}^n \in \hat{\mathcal{Y}}_{src},\\
target & \text{if}\ \mathbf{y}^n \in \hat{\mathcal{Y}}_{tgt},\\
generated & \text{if}\ \mathbf{y}^n \in \mathcal{Y}_{\pi_\theta},\\
\end{cases}
\\
\mathcal{Y}_{\pi_\theta}=\{\mathbf{y}^n\sim \pi_\theta(.|\mathbf{x}^n,.)\}_n, \mathbf{x}^n\in \mathcal{X}_{tgt}~,\label{Eq.DC}
\end{split}
\end{equation}
where $N$ is the number of sentences, $\phi$ is the model parameter of DC, $\hat{\mathcal{Y}}_{src}$ denotes sentences from the source domain, $\hat{\mathcal{Y}}_{tgt}$ denotes sentences from the target domain, and $\mathcal{Y}_{\pi_\theta}$ denotes sentences generated from the captioner with policy $\pi_\theta$ given target domain images $\mathcal{X}_{tgt}$.

For MC, the goal is to classify a image-sentence pair into paired, unpaired, and generated data. This can also be formulated as a supervised classification training objective as follows,
\vspace{-3mm}
\begin{equation} \small 
\begin{split}
\mathcal{L}_{m}(\eta)=-\sum_{n=1}^{N} \log C_m(l^n|\mathbf{x}^n, \mathbf{y}^n;\eta)~,\\
l^n=
\begin{cases}
paired & \text{if}\ (\mathbf{x}^n,\mathbf{y}^n) \in \mathcal{P}_{src}~,\\
unpaired & \text{if}\ (\mathbf{x}^n,\mathbf{y}^n) \in \acute{\mathcal{P}}_{src}~,\\
generated & \text{if}\ (\mathbf{x}^n,\mathbf{y}^n) \in \mathcal{P}_{gen}~,
\end{cases}
\\
\acute{\mathcal{P}}_{src} = \{(\mathbf{x}^i\in \mathcal{X}_{src},\hat{\mathbf{y}}^j\in \hat{\mathcal{Y}}_{src});i\neq j\}~,\\
\mathcal{P}_{gen} = \{(\mathbf{x}\in \mathcal{X}_{src},\mathbf{y}\in \mathcal{Y}_{\pi_\theta})\}~,\label{Eq.MC}
\end{split}
\end{equation}
where $\eta$ is the model parameter of MC, $\mathcal{P}_{src}$ is the paired data from the source domain, $\acute{\mathcal{P}}_{src}$ is the unpaired data intentionally collected randomly by shuffling images and sentences in the source domain, and $\mathcal{P}_{gen}$ is the source-image-generated-sentence pairs.
\vspace{-0mm}
\begin{algorithm}[t]
\small
  \caption{Adversarial Training Procedure}\label{algo:overall}

 \SetKwInput{Require}{Require}
 \Indm
 \Require{captioner $\pi_\theta$, domain critic $C_d$, multi-modal critic $C_m$, an empty set for generated sentences $\mathcal{Y}_{\pi_\theta}$, and an empty set for paired image-generated-sentence $\mathcal{P}_{gen}$;}
 
 \KwIn{sentences $\hat{\mathcal{Y}}_{src}$, image-sentence pairs $\mathcal{P}_{src}$, unpaired data $\acute{\mathcal{P}}_{src}$ in source domain; sentences $\hat{\mathcal{Y}}_{tgt}$, images $\mathcal{X}_{tgt}$ in target domain;}
 \Indp
  Pre-train $\pi_\theta$ on $\mathcal{P}_{src}$ using Eq.~\ref{Eq.ori}\;
  \While{$\theta$ has not converged}
 {
  \For{$i=0,...,N_{c}$}
  {
    $\mathcal{Y}_{\pi_\theta} \leftarrow \{\mathbf{y}\},
    \textrm{where } \mathbf{y}\sim \pi_{\theta}(\cdot|\mathbf{x},\cdot) \textrm{ and } \mathbf{x} \sim \mathcal{X}_{tgt}$\;
    Compute $g_d = \nabla_\phi \mathcal{L}_{d}(\phi)$ using Eq.~\ref{Eq.DC}\;
    Adam update of $\phi$ for $C_d$ using $g_d$\;
    $\mathcal{Y}_{\pi_\theta} \leftarrow \{\mathbf{y}\},
    \textrm{where } \mathbf{y}\sim \pi_{\theta}(\cdot|\mathbf{x},\cdot) \textrm{ and } \mathbf{x} \sim \mathcal{X}_{src}$\;
    $\mathcal{P}_{gen} \leftarrow \{(\mathbf{x},\mathbf{y})\}$\;
    Compute $g_m = \nabla_{\eta} \mathcal{L}_{m}(\eta)$ using Eq.~\ref{Eq.MC}\;
    Adam update of $\eta$ for $C_m$ using $g_m$\;

  }
  \For{$i=0,...,N_{g}$}
  {
  $\mathcal{Y}_{\pi_\theta} \leftarrow \{\mathbf{y}\},
    \textrm{where } \mathbf{y}\sim \pi_{\theta}(\cdot|\mathbf{x},\cdot) \textrm{ and } \mathbf{x} \sim \mathcal{X}_{tgt}$\;
  $\mathcal{P}_{gen} \leftarrow  \{ (\mathbf{x},\mathbf{y}) \}$\;
  \For{$t=1,...,T$}
  	{
    Compute $Q((\mathbf{x},\mathbf{y}_{t-1}), y_t)$ with Monte Carlo rollouts, using Eq.~\ref{Eq.RO}\;
    }
   Compute $g_\theta=\nabla_\theta{J(\theta)}$ using Eq.~\ref{Eq.dJtotal}\;
   Adam update of $\theta$\ using $g_\theta$;
  }
 }
\vspace{-1mm}
\end{algorithm}

\vspace{-2mm}
\subsection{Adversarial Training}
\vspace{-2mm}

Our cross-domain image captioning system is summarized in Fig.~\ref{fig:model}.
Both captioner $\pi_\theta$ and critics $C_d$ and $C_m$ learn together by pursuing competing goals as described below.
Given $\mathbf{x}$, the captioner $\pi_\theta$ generates a sentence $\mathbf{y}$. It would prefer the sentence to have large reward $R(\mathbf{y}|.)$, which implies large values of $C_d(target|\mathbf{y})$ and $C_m(paired|\mathbf{x}.\mathbf{y})$.
In contrast, the critics would prefer large values of $C_d(generated|\mathbf{y})$ and $C_m(generated|\mathbf{x},\mathbf{y})$, which implies small values of $C_d(target|\mathbf{y})$ and $C_m(paired|\mathbf{x}.\mathbf{y})$. We propose a novel adversarial training procedure to iteratively updating the captioner and critics in Algorithm~\ref{algo:overall}. In short, we first pre-train the captioner using cross-entropy loss on source domain data. Then, we iteratively update the captioner and critics with a ratio of $N_{g}:N_{c}$, where the critics are updated more often than captioner (i.e., $N_{g}<N_{c}$).

\begin{figure*}[t!]\vspace{-9mm}
\begin{center}
\includegraphics[width=0.85\textwidth]{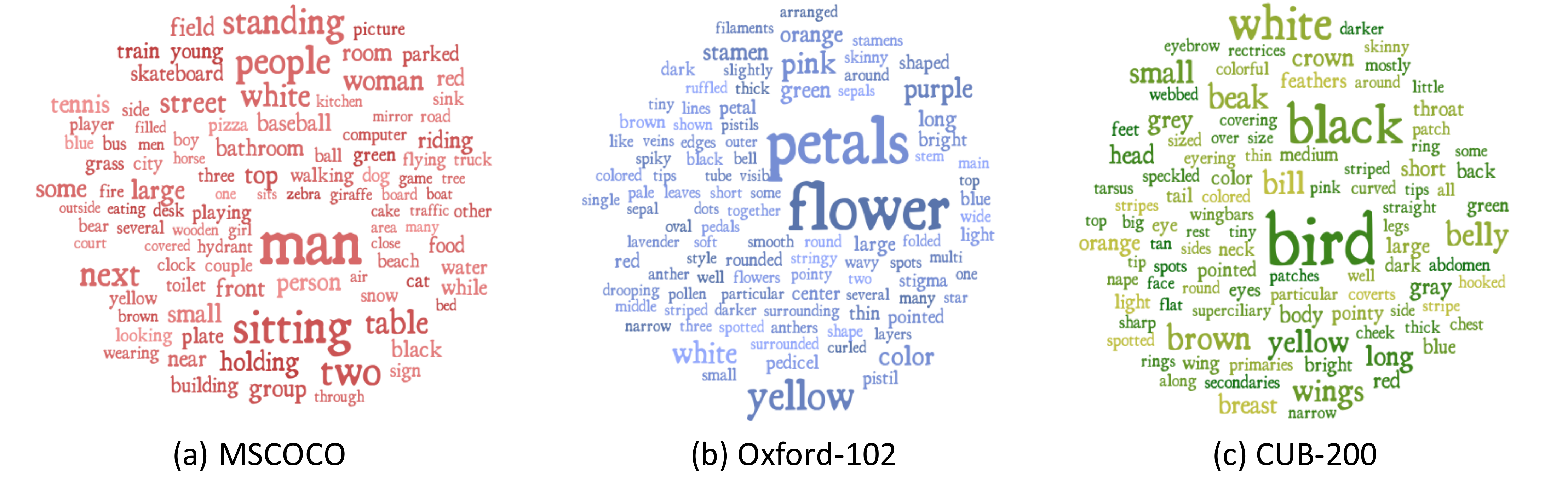}
\end{center}
\vspace{-5mm}
\caption{\small Word clouds for testing set of (a) MSCOCO, (b) Oxford-102, (c) CUB-200, where font size indicates the frequency of words.
}\label{fig:cloud}
\vspace{-3mm}
\end{figure*}

\subsection{Critic-based Planning}



The quality of a generated word $y_t$ is typically measure by the policy network $\pi(y_t|\cdot)$. For cross-domain captioning, the learned critics can also be used to measure the quality of $y_t$ by computing $Q((\mathbf{x}, \mathbf{y}_{t-1}),y_t)$ using Eq.~\ref{Eq.RO}. Here, $Q$ is an expected value that models the randomness of future words, so we call our method "critic-based planning". Critic-based planning takes advantage of both the learned policy network as well as the critics.
By default, we select $y^*_t=\arg\max_y\pi_\theta(y|\cdot)$ as the generated word. However, when the difference between the maximum probability and the second largest probability of $\pi_\theta(\cdot)$ is below a threshold $\Gamma$ 
(where the selection of $y^*_t$ is ambiguous), we take the top $J$ words $\{y_t^j\}_{j=1}^J$ according to $\pi(y|\cdot)$ and evaluate $Q((\mathbf{x}, \mathbf{y}_{t-1}),y_t^j)$ for all $j$. Then, we select the word with the highest $Q$ value as the generated word. Note that the sentences generated via critic-based planning can be exactly the same as greedy search. Our critic-based planning method obtain further performance improvement typically on dataset with large domain shift (e.g., CUB-200 and Oxford-102).







\begin{table*}[t!]\vspace{-5mm}
\centering\scriptsize
\caption{\small{Results of adaptation across four target domain datasets. Source (MSCOCO) Pre-trained and DCC are two baseline methods. Fine-tuning with paired data in target domain serves as the upper bound performance of our CNN-RNN captioner.}}\label{table:main}
\label{transfer-main}
\vspace{-2mm}
\scalebox{1.15}{
\begin{tabular}{@{}lccccccccc@{}}
\toprule
\multicolumn{1}{c}{Method}    & Target (test) & Bleu-1 & Bleu-2 & Bleu-3 & Bleu-4  & Meteor & ROUGE & CIDEr & SPICE \\ \toprule
Source Pre-trained                & CUB-200   & 50.8   & 28.3   &  13.9   & 6.1 & 12.9 & 33 & 3 & 4.6  \\
DCC   & CUB-200   & 68.6 & 47.3	 &	31.4 &	21.4 &	23.8 &	46.4 &	11.9 & 11.1 \\
Ours              & CUB-200   & \bf{91.4}   & \bf{73.1}   &  \bf{51.9}   & \bf{32.8} & \bf{27.6} & \bf{58.6} & \bf{24.8}  & \bf{13.2}   \\
\rowcolor{mygray}
Fine-tuning                   & CUB-200   & 91.3 & 80.2   &  69.2  &  59    & 36.1 & 69.7 & 61.1 & 17.9   \\ \midrule
Source Pre-trained                 & Oxford-102   & 48.3   &  21.6 & 6.2 &	1.3	& 10.5 & 25.8 &	3.1 & 4.4  \\
DCC  			& Oxford-102 & 51 & 33.8 & 24.1 & 16.7 & 21.5 &	38.3 & 6 & 9.8 \\
Ours             & Oxford-102   & {\bf 85.6}	& {\bf 76.9} & {\bf 67.4} &	{\bf 60.5} &	{\bf 36.4} &	{\bf 72.1} &	{\bf 29.3} & \bf{17.9}   \\
\rowcolor{mygray}
Fine-tuning                   & Oxford-102   & 87.5 &	80.1 &	72.8 &	66.3	& 40 &	75.6 &	36.3 & 18.5   \\ \midrule
Source Pre-trained                  & TGIF      &41.6&	23.3&	12.6&	7&	12.7&	32.7&	14.7 & 8.5  \\
DCC  			& TGIF & 34.6 &	17.5 & 9.3 & 4.1 &	11.8	& 29.5	& 7.1 & 7.3 \\
Ours            & TGIF   &  {\bf 47.5}&	{\bf 29.2}&	{\bf 17.9}&	{\bf 10.3}&	{\bf 14.5}&	{\bf 37}&	{\bf 22.2}  & \bf{10.6} \\
\rowcolor{mygray}
Fine-tuning                     & TGIF      & 51.1&	32.2&	20.2&	11.8&	16.2&	39.2&	29.8  & 12.1\\
\midrule
Source Pre-trained                  & Flickr30k &57.3	&36.2	&21.9&	13.3&	15.1&	38.8&	25.3  & 8.6  \\
DCC 		 	& Flickr30k & 54.3 & 34.6 & 21.8 & 13.8 & 16.1 & 38.8 & 27.7 & 9.7 \\
Ours             & Flickr30k & {\bf 62.1}&	{\bf 41.7}&	{\bf 27.6}&	{\bf 17.9}&	{\bf 16.7}&	{\bf 42.1}&	{\bf 32.6} & \bf{9.9}  \\
\rowcolor{mygray}
Fine-tuning                   & Flickr30k & 59.8&	41&	27.5&	18.3&	18&	42.9&	35.9 & 11.5  \\ 
 \bottomrule
\end{tabular}
}
\end{table*}

\vspace{-4mm}
\section{Experiments}
\vspace{-2mm}
We perform extensive evaluations on a number of popular datasets. For all experiments, we use 
MSCOCO~\cite{lin2014microsoft} as the source dataset and CUB-200~\cite{wah2011caltech}, Oxford-102~\cite{Nilsback08}, TGIF~\cite{Li:CVPR16}, and Flickr30k~\cite{young2014image} as target datasets. We show that our method generalizes to datasets with large domain shift (CUB-200 and Oxford-102) and datasets with regular domain shift (Flickr30k and TGIF). We also show that critic-based planning can further improve performance during inference on datasets with large domain shift. Finally, we conduct an ablation study on Flickr30k to show the contribution of different components.


\subsection{Implementation details}
\noindent\textbf{Data preprocessing.} 
For source domain dataset, we select the MSCOCO training split from \cite{karpathy2015deep} which contains $113,287$ images, along with 5 captions each. We prune the vocabulary by dropping words with frequency less than $5$, resulting in $10,066$ words including special Begin-Of-Sentence (BOS) and End-Of-Sentence (EOS) tokens. We use the same vocabulary in all experiments.
For target domain datasets, we remove the training sentences containing out-of-vocabulary words (see Supplementary for detailed statistics).

\noindent\textbf{Pre-training details.} The architecture of our captioner is a CNN-LSTM with hidden dimension $512$. The image features are extracted using the pre-trained Resnet-101~\cite{ResNet15} and the sentences are represented as one-hot encoding. We first pre-train the captioner on source domain dataset via cross entropy objective using ADAM optimizer~\cite{KingmaB14} with learning rate $5 \times 10^{-4}$. We apply learning rate decay with a factor of $0.8$ every three epoches. To further improve the performance, we use schedule sampling~\cite{bengio2015scheduled} to mitigate the exposure bias. The best model is selected according to the validation performance and serve as the initial model for adversarial training.

\noindent\textbf{Adversarial training details.}
We train the captioner and critics using ADAM optimizer~\cite{KingmaB14} with learning rate of $5 \times 10^{-5}$. We apply dropout in training phase to prevent over-fitting, which also served as input noise similar to \cite{isola2016image}. In Monte Carlo rollout, the model samples words until the EOS token under the current policy for $K=3$ times. These $K$ sentences are them fed to the critics for estimating the state-action value $Q(\cdot)$. Both critics are trained from scratch using the standard classification objective.


\subsection{Experimental Results}
We first pre-train the captioner on MSCOCO training set.
Next, we update the captioner by adversarial training procedure with unpaired data from the training set in target domains.  
Finally, we evaluate our method on four target domain datasets, representing different levels of domain shift.


\noindent\textbf{Baseline.}
We re-implement Deep Compositional Captioner (referred to as DCC)~\cite{hendricks16cvpr} as our baseline method.
DCC consists of a lexical classifier and a language model. The former is a CNN model trained to predict semantic attributes and the latter is an LSTM model trained on unpaired text. In the end, the overall DCC model combines both models with a linear layer trained on paired image-caption data. For fair comparison, we apply the following settings on DCC, where the lexical classifier is a ResNet-101 model and the language model is trained on target domain sentences. 
\Paul{Note that the ResNet-101 is fine-tuned with 471 visual concepts (pre-defined in \cite{hendricks16cvpr}) extracted from captions.}
Finally, we use source domain image-caption pairs to fine-tune DCC. 
We also fine-tune a pre-trained source domain model directly on paired training data in the target domain (referred to as Fine-tuning). Ideally, this serves as the upper bound~\footnote{We find that the model directly trained on all paired data in target domain performs worse than fine-tuning. Please see Supplementary for details. } of our experiments.

\ignore{
\begin{table}[h]
\caption{Number of unpaired training captions for inductive and transductive domain adaptation.}
\label{Table:dataset-stat}
\centering
\begin{tabular}{@{}ccc@{}}
\toprule
\multirow{2}{*}{Dataset} & \multicolumn{2}{c}{\# captions} \\ \cmidrule(l){2-3} 
                         & Inductive          & Transductive          \\ \midrule
CUB-200                  & 25,926             & 63,787                \\ 
Oxford-102               & 22,716             & 27,512                \\ 
Flickr30k                & 117,664            & 121,763               \\ 
TGIF                     & 65,526             & 93,718                \\ \bottomrule
\end{tabular}
\end{table}
}

We further categorize three kinds of domain shift between MSCOCO and other target datasets, namely general v.s. fine-grained descriptions, difference in verb usage and subtle difference in sentence style.

\noindent\textbf{General v.s. fine-grained descriptions.}
The large domain shift between MSCOCO and CUB-200/Oxford-102 suggests that it is the most challenging domain adaptation scenario. In CUB-200/Oxford-102, descriptions give detailed expressions of attributes such as beak of a bird or stamen of a flower. In contrast, in MSCOCO, descriptions usually are about the main scene and character. 
We illustrate the differences at word-level distribution among MSCOCO, CUB-200, and Oxford-102 using Venn-style word clouds~\cite{coppersmith2014dynamic} (see Fig.~\ref{fig:cloud}~\footnote{Visualization generated using http://worditout.com/.}). 

\begin{figure*}[h]\vspace{-7mm}
\begin{center}
\includegraphics[width=1\textwidth]{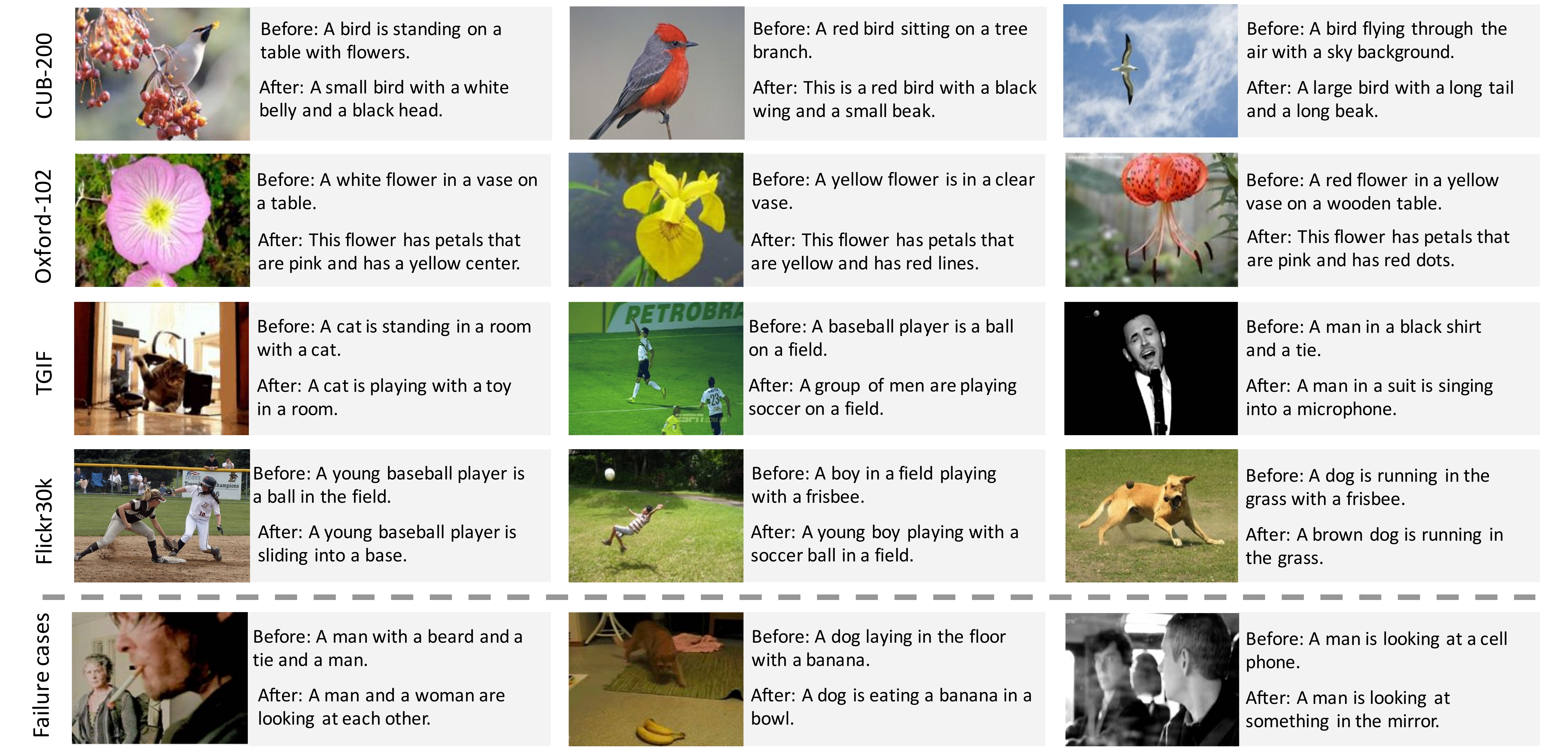}
\end{center}
\vspace{-7mm}
\caption{\small Examples of captions before and after domain adaptation for all four target domain datasets. \Paul{The last row demonstrates the failure cases, where the generated captions do not accurately describe the images. }
}\label{fig:typ_example}
\end{figure*}

On the top two rows of Fig.~\ref{fig:typ_example} show that our model can describe birds and flowers in detailed and also the appearance of fine-grained object attributes.
In the top two blocks of Table~\ref{table:main}, our method outperforms DCC and Source Pre-trained models by a considerable margin for all evaluation metrics.

\noindent\textbf{Difference in verb usage.}
Next, we move towards the verb usage difference between the source and target domains.
According to \cite{Li:CVPR16}, there are more motion verbs (30\% in TGIF vs. 19\% in MSCOCO) such as dance and shake, and more facial expressions in TGIF, while verbs in MSCOCO are mostly static ones such as stand and sit. Examples in Fig.~\ref{fig:typ_example} show that our model can accurately describe human activities or object interactions. On the third panel of Table~\ref{table:main}, our method also significantly improves over Source Pre-trained and DCC models.

\vspace{-1mm}
\noindent\textbf{Subtle difference in sentence style.}
In order to test the generalizability of our method, we conduct an experiment using similar dataset (i.e. Flickr30k) as target domain. In the bottom block of Table~\ref{table:main}, our method also offers a noticeable improvement. 
\Paul{In addition, we reverse the route of adaptation (i.e. from Flickr30k to MSCOCO). Our method (CIDEr 38.2\%, SPICE 8.9\%) also improves over source pre-trained model (CIDEr 27.3\%, SPICE 7.6\%).}
To sum up, our method shows great potentials for unsupervised domain adaptation across datasets regardless of regular or large domain shift. 

\begin{figure*}[h]\vspace{-5mm}
\begin{center}
\includegraphics[width=0.9\textwidth]{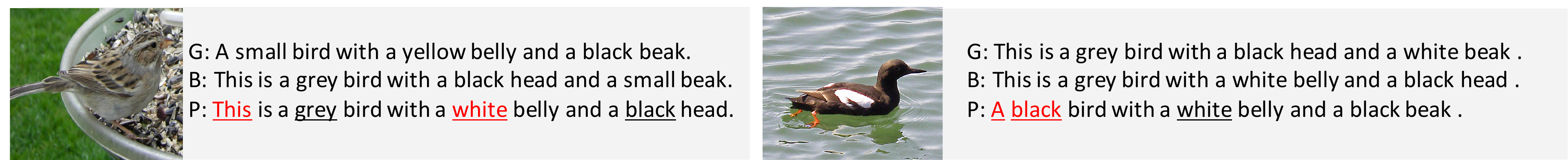}
\end{center}
\vspace{-4mm}
\caption{\small{Results of critic-based planning. G stands for greedy search, B for beam search, and P for critic-based planning. The underlined words denote that the difference between the maximum probability and the second largest probability of $\pi$ is lower than $\Gamma$ (selected by critic). When critic-based planning does not choose the word with maximum probability of $\pi$, the word is colored in red.}
}\label{fig:planning_example}
\end{figure*}
\vspace{-1mm}
\noindent\textbf{Critic-based planning.}
Instead of directly generating the word $y_{t}$ from policy network $\pi(y_{t}|.)$, we take the advantage of its adversary, critics, during the inference. The results is shown in Table~\ref{table:planning}. The threshold $\Gamma$ is set to 0.15 in CUB-200 and to 0.1 in Oxford-102. In every time-step, we choose top $J=2$ words according to $\pi_{\theta}(.)$. Out of 16.2\% and 9.4\% of words are determined by the critics in CUB-200, and Oxford-102, respectively.
Compared to greedy search, critic-based planning can achieve better performance in many evaluation metrics, especially in datasets with large domain shift from the source domain dataset (e.g., CUB-200 and Oxford-102). Compared to beam search with beam size 2, critic-based planning also typically gets a higher performance. Beam search method generates the words only depending on captioner itself, while critic-based planning method acquires a different point of view from the critics. 
\Paul{
For the case of regular domain shift (e.g., TGIF and Oxford-102), critic-based planning achieves comparable performance with beam search and greedy search.
}
Some impressive examples are shown in Fig.~\ref{fig:planning_example}.

\begin{table}[h]\vspace{-6mm}
\centering\scriptsize
\caption{\small{Results of proposed critic-based planning compared with greedy search and beam search.}}
\label{table:planning}
\vspace{-2mm}
\scalebox{1.1}{
\begin{tabular}{@{}llllll@{}}
\toprule
\multicolumn{1}{c}{Method} & Bleu-4        & Meteor        & ROUGE         & CIDEr-D     \\ \midrule
\multicolumn{5}{c}{MSCOCO $\,\to\,$ CUB-200}                             \\ \midrule
Greedy Search   & 32.8          & 27.6          & 58.6          & 24.8          \\ 
Beam Search     & 33.1          & 27.5          & 58.3          & 26.2          \\ 
Planning        & \bf{35.2}     & 27.4          & 58.5          & \bf{29.3}          \\ \midrule
\multicolumn{5}{c}{MSCOCO $\,\to\,$ Oxford-102}                               \\ \midrule
Greedy Search   & 60.5          & 36.4          & 72.1          & \bf{29.3} \\ 
Beam Search     & 60.3          & 36.3          & 72            & 28.3          \\ 
Planning        & \bf{62.4} & \bf{36.6} & \bf{72.6} & 24.9  
\\ \midrule
\multicolumn{5}{c}{MSCOCO $\,\to\,$ TGIF}                               \\ \midrule
Greedy Search  & 10.3          & 14.5          & 37          & 22.2         \\ 
Beam Search     & 10.5          & 14.2          & 36.7          & 22.6          \\ 
Planning        & 10.3     & 14.4          & 37          & 21.9          \\ \midrule
\multicolumn{5}{c}{MSCOCO $\,\to\,$ Flickr30k}                               \\ \midrule
Greedy Search   & 17.5          & 16.4          & 41.9          & 32.2          \\ 
Beam Search     & 18.2          & 16.4          & 42.1          & 33.3          \\ 
Planning        & 17.3     & 16.5          & 41.7          & 32.3          \\ \bottomrule

\end{tabular}
}
\end{table}
\vspace{-3mm}
\subsection{Ablation Study}
\vspace{-2mm}
We have proposed an adversarial training procedure with two critic models: Multi-modal Critic (MC) and Domain Critic (DC). In order to analyze the effectiveness of these two critics, we do ablation comparison with either one and both. Table~\ref{table:ablation} shows that using MC only is insufficient since MC is not aware of the sentence style in target domain. On the other hand, using DC only contributes significantly. Finally, combining both MC and DC achieves the best performance for all evaluation metrics. We argue that both MC and DC are vital for cross-domain image captioning. 

\begin{table}[h]\vspace{-3mm}
\centering\scriptsize
\caption{\small{Ablation study for two critic models on Flickr30k. MC: Multi-modal Critic, DC: Domain Critic.}}
\label{table:ablation}
\vspace{-2mm}
\scalebox{1.1}{
\begin{tabular}{@{}lccccccc@{}}
\toprule
\multicolumn{1}{c}{Method} & Bleu-4  & Meteor & ROUGE & CIDEr-D \\ \midrule
Source Pre-trained   & 13.3 & 15.1  &  38.8   & 25.3   \\
+MC                  & 13.7	& 15.2 & 38.8 &	25.9   \\
+DC                  & 17.6 & 16.3 & 41.4 &	32.1   \\
+MC+DC               & \bf{17.9} & \bf{16.7} & \bf{42.1} &	\bf{32.6}      \\ 
\bottomrule
\end{tabular}
}
\end{table}

\vspace{-7mm}
\section{Conclusion}
\vspace{-2mm}
We propose a novel adversarial training procedure (captioner v.s. critics) for cross-domain image captioning. A novel critic-based planning method is naturally introduced to further improve the caption generation process in testing.
Our method consistently outperforms baseline methods on four challenging target domain datasets (two with large domain shift and two with regular domain shift). In the future, we would like to improve the flexibility of our method by combining multiple critics in a plug-and-play fashion.

\small\textbf{Acknowledgement}
We thank Microsoft Research Asia, MediaTek and MoST 106-3114-E-007-004 for their support. We also thank Kuo-Hao Zeng and Shao-Pin Chang for useful feedbacks during internal review.

\small{
\bibliographystyle{ieee}

}

\end{document}